\documentclass[runningheads]{llncs}
\usepackage{graphicx}

\usepackage{subfig}
\usepackage{amsmath,amssymb} 
\usepackage{color}
\usepackage{cite}

\begin{document}
	
	\title{Random Temporal Skipping for Multirate Video Analysis} 
	\titlerunning{Random Temporal Skipping} 
	
	
	\author{Yi Zhu\inst{1} \and Shawn Newsam\inst{1}}
	%
	
	\authorrunning{Y. Zhu et al.} 
	
	
	\institute{University of California at Merced, Merced CA 95343, USA 
		\email{\{yzhu25,snewsam\}@ucmerced.edu}}
	
	\maketitle
	
	\begin{abstract}
		Current state-of-the-art approaches to video understanding adopt temporal jittering to simulate analyzing the video at varying frame rates. However, this does not work well for multirate videos, in which actions or subactions occur at different speeds. The frame sampling rate should vary in accordance with the different motion speeds. In this work, we propose a simple yet effective strategy, termed random temporal skipping, to address this situation. This strategy effectively handles multirate videos by randomizing the sampling rate during training. It is an exhaustive approach, which can potentially cover all motion speed variations. Furthermore, due to the large temporal skipping, our network can see video clips that originally cover over 100 frames. Such a time range is enough to analyze most actions/events. We also introduce an occlusion-aware optical flow learning method that generates improved motion maps for human action recognition. Our framework is end-to-end trainable, runs in real-time, and achieves state-of-the-art performance on six widely adopted video benchmarks.
		
	\end{abstract}
\section{Introduction}
\label{sec:intro}

Significant progress has been made in video analysis during the last five years, including content-based video search, anomaly detection, human action recognition, object tracking and autonomous driving. Take human action recognition as an example. The performance on the challenging UCF101 dataset \cite{ucf101} was only $43.9\%$ reported in the original; it now is $98.0\%$. 
Such great improvement is attributed to several factors, such as more complicated models (e.g., deep learning \cite{I3D_Carreira_cvpr17}), larger datasets (e.g., Kinetics \cite{kinetics}), better temporal analysis (e.g., two-stream networks \cite{twostream2014,hidden_zhu_18}), etc. 

\begin{figure*}[t]
	\centering
	\includegraphics[width= 1.0\linewidth]{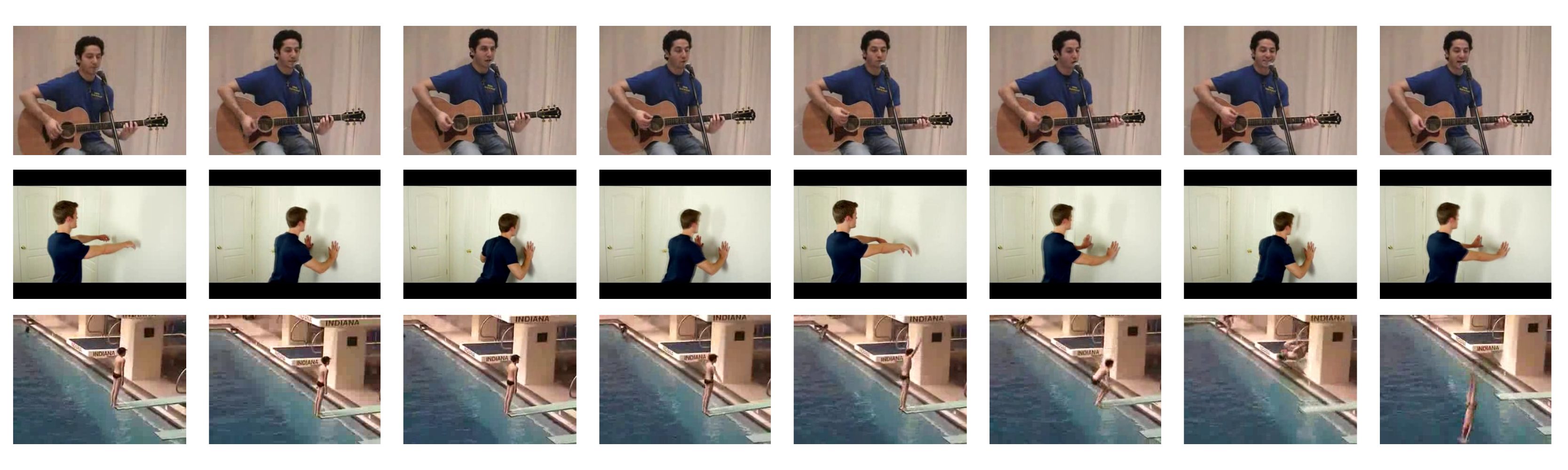}
	\caption{Sample video frames of three actions: (a) playingguitar (b) wallpushup and (c) diving. (a) No temporal analysis is needed because context information dominates. (b) Temporal analysis would be helpful due to the regular movement pattern. (c) Only the last four frames have fast motion, so multirate temporal analysis is needed.}
	\label{fig:idea}
\end{figure*}

However, there has been little work on varying frame-rate video analysis. For simplicity, we denote varying frame-rate as multirate throughout the paper. For real-world video applications, multirate handling is crucial. For surveillance video monitoring, communication package drops occur frequently due to bad internet connections. We may miss a chunk of frames, or miss the partial content of the frames. For activity/event analysis, the videos are multirate in nature. People may perform the same action at different speeds. For video generation, we may manually interpolate frames or sample frames depending on the application. For the scenarios mentioned above, models pre-trained on fixed frame-rate videos may not generalize well to multirate ones. As shown in Figure \ref{fig:idea}, for the action diving, there is no apparent motion in the first four frames, but fast motion exists in the last four frames. Dense sampling of every frame is redundant and results in large computational cost, while sparse sampling will lose information when fast motion occurs. 

There are many ways to model the temporal information in a video, including trajectories \cite{idtfWang2013}, optical flow \cite{twostream2014}, temporal convolution \cite{longTemporalConv2016}, 3D CNNs \cite{c3d2015} and recurrent neural networks (RNNs) \cite{beyondshort2015}. However, none of these methods can directly handle multirate videos. Usually these methods need a fixed length input (a video clip) with a fixed sampling rate. 
A straightforward extension therefore is to train multiple such models, each corresponding to a different fixed frame-rate. This is similar to using image pyramids to handle the multi-scale problem in image analysis. But it is computational infeasible to train models for all the frame-rates. 
And, once the frame-rate differs, the system's performance may drop dramatically. Hence, it would be more desirable to use one model to handle multiple frame-rates.

In this work, we focus on human action recognition because action is closely related to frame-rate. Specifically, our contributions include the following. First, we propose a random temporal skipping strategy for effective multirate video analysis. It can simulate various motion speeds for better action modeling, and makes the training more robust. Second, we introduce an occlusion-aware optical flow learning method to generate better motion maps for human action recognition. Third, we adopt the ``segment'' idea \cite{TSN2016,diba_tle_2016} to reason about the temporal information of the entire video. By combining the local random skipping and global segments, our framework achieves state-of-the-art results on six large-scale video benchmarks. In addition, our model is robust under dramatic frame-rate changes, a scenario in which the previous best performing methods \cite{TSN2016,diba_tle_2016,I3D_Carreira_cvpr17} fail.

\section{Related Work}

There is a large body of literature on video human action recognition. Here, we review only the most related work. 

\noindent \textbf{Deep learning for action recognition} 
Initially, traditional handcrafted features such as Improved Dense Trajectories (IDT) \cite{idtfWang2013} dominated the field of video analysis for several years. Despite their superior performance, IDT and its improvements \cite{tddwang2015} are computationally formidable for real-time applications. CNNs \cite{KarpathyCVPR14,c3d2015}, which are often several orders of magnitude faster than IDTs, performed much worse than IDTs in the beginning. 
Later on, two-stream CNNs \cite{twostream2014} addressed this problem by pre-computing optical flow and training a separate CNN to encode the pre-computed optical flow. This additional stream (a.k.a., the temporal stream) significantly improved the accuracy of CNNs and finally allowed them to outperform IDTs on several benchmark action recognition datasets. These accuracy improvements indicate the importance of temporal motion information for action recognition.

\noindent \textbf{Modeling temporal information} 
However, compared to the CNN, the optical flow calculation is computationally expensive. It is thus the major speed bottleneck of the current two-stream approaches. There have been recent attempts to better model the temporal information. Tran et al.  \cite{c3d2015} pre-trained a deep 3D CNN network on a large-scale dataset, and use it as a general spatiotemporal feature extractor. The features generalize well to several tasks but are inferior to two-stream approaches. Ng et al. \cite{beyondshort2015} reduced the dimension of each frame/clip using a CNN and aggregated frame-level information using Long Short Term Memory (LSTM) networks. Varol et al.  \cite{longTemporalConv2016} proposed to reduce the size of each frame and use longer clips (e.g., 60 vs 16 frames) as inputs. They managed to gain significant accuracy improvements compared to shorter clips with the same spatial size. Wang et al.  \cite{TSN2016} experimented with sparse sampling and jointly trained on the sparsely sampled frames/clips. In this way, they incorporate more temporal information while preserving the spatial resolution. Recent approaches \cite{diba_tle_2016,dovf_lan_2017} have evolved to end-to-end learning and are currently the best at incorporating global temporal information. However, none of them handle multirate video analysis effectively. 

\noindent \textbf{Multi-rate video analysis} To handle multirate videos, there are two widely adopted approaches. One is to train multiple models, each of them corresponding to a different fixed frame-rate. This is similar to using image pyramids to handle the multi-scale problem in image analysis. The other is to generate sliding windows of different lengths for each video (a.k.a, temporal jittering), with the hope of capturing temporal invariance. However, neither of these approaches is exhaustive, and they are both computationally intensive. 
\cite{multirate_zhu_cvpr2017} is the most similar work to ours since they deal with motion speed variance. However, our work differs in several aspects. First, we aim to explicitly learn the transitions between frames while \cite{multirate_zhu_cvpr2017} uses past and future neighboring video clips as the temporal context, and reconstruct the two temporal transitions. Their objective is considerably harder to optimize, which may lead to sub-optimal solutions. Second, our random skipping strategy is easy to implement without computational overhead whereas the image reconstruction of \cite{multirate_zhu_cvpr2017} will lead to significant computational burden. Third, their proposed multirate gated recurrent unit only works in RNNs, while our strategy is generally applicable.

In conclusion, to overcome the challenge that CNNs are incapable of capturing temporal information, we propose an occlusion-aware CNN to estimate accurate motion information for action recognition. To handle multirate video analysis, we introduce random temporal skipping to both capture short motion transitions and long temporal reasoning. Our framework is fast (real-time), end-to-end optimized and invariant to frame-rate. 

\section{Approach}

There are two limitations to existing temporal modeling approaches: they require a fixed length input and a fixed sampling rate. For example, we usually adopt 16 frames to compute IDT and C3D features, 10 frames to compute optical flow for two-stream networks, and 30 frames for LSTM. These short durations do not allow reasoning on the entire video. In addition, a fixed sampling rate will either result in redundant information during slow movement or the loss of information during fast movement. The frame sampling rate should vary in accordance with different motion speeds. 
Hence, we propose random temporal skipping.

\subsection{Random Temporal Skipping}
In this section, we introduce random temporal skipping and illustrate its difference to traditional sliding window (fixed frame-rate) approaches. For easier understanding, we do not use temporal segments here. 

Consider a video $V$ with a total of $T$ frames $[v_{1}, v_{2}, \dots, v_{T}]$. In the situation of single-rate analysis, we randomly sample fixed length video clips from an entire video for training. Suppose the fixed length is $N$, then the input to our model will be a sequence of frames as 
\begin{equation}
	[v_{t}, v_{t+1}, \cdots, v_{t+N}].
\end{equation}
In order to learn a frame-rate invariant model, a straightforward way is using a sliding window. The process can be done either offline or online. The idea is to generate fixed length video clips with different temporal strides, thus covering more video frames. Much literature adopts such a strategy as data augmentation. Suppose we have a temporal stride of $\tau$. The input now will be  
\begin{equation}
	[v_{t}, v_{t+\tau}, \cdots, v_{t+N\tau}].
\end{equation}
As shown in Figure \ref{fig:idea}, a fixed sampling strategy does not work well for multirate videos. A single $\tau$ can not cover all temporal variations. The frame sampling rate should vary in accordance with different motion speeds. Motivated by this observation, we propose random temporal skipping. Instead of using a fixed temporal stride $\tau$, we allow it vary randomly. The input now will be 
\begin{equation}
	[v_{t}, v_{t+\tau_{1}}, \cdots, v_{t+\tau_{1} + \tau_{2} + \cdots + \tau_{N}}].
	\label{eq:skip}
\end{equation}
Here, $\tau_{n}$, $n=1,2,\cdots,N$ are randomly sampled within the range of [0, \textit{maxStride}]. \textit{maxStride} is a threshold value indicating the maximum distance we can skip in the temporal domain. Our proposed random temporal skipping represents an exhaustive solution. Given unlimited training iterations, we can model all possible combinations of motion speed, thus leading to the learning of frame-rate invariant features. In addition, this strategy can be easily integrated into existing frameworks with any model, and can be done on-the-fly during training. 

\begin{figure*}[t]
	\centering
	\includegraphics[width= 1.0\linewidth]{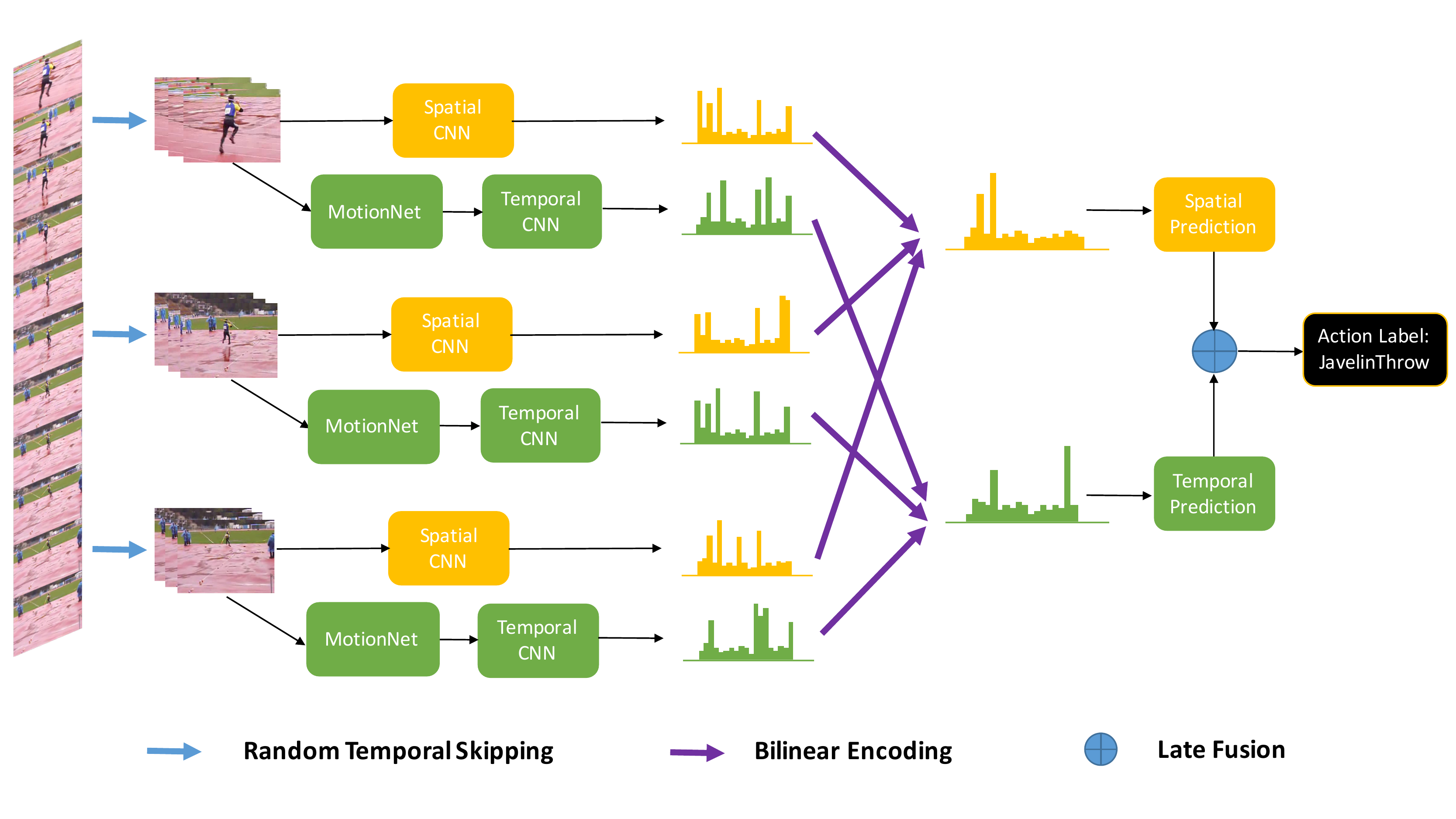}
	\caption{Overview of our proposed framework. Our contributions are three fold: (a) random temporal skipping for temporal data augmentation; (b) occlusion-aware MotionNet for better motion representation learning; (c) compact bilinear encoding for longer temporal context. }
	\label{fig:network}
\end{figure*}

\subsection{Two-Stream Network Details}
Since two-stream networks are the state-of-the-art \cite{TSN2016,I3D_Carreira_cvpr17} for several video benchmarks, we also build a two-stream model but with significant modifications. In this section, we first briefly recall temporal segment network (TSN) to illustrate the idea of segments. Then we describe our newly designed spatial and temporal streams, respectively. 

\noindent \textbf{Temporal segment network} 
With the goal of capturing long-range temporal structure for improved action recognition, Wang et al. proposed TSN \cite{TSN2016} with a sparse sampling strategy. This allows an entire video to be analyzed with reasonable computational costs. TSN first divides a video evenly into three segments and one short snippet is randomly selected from each segment. Two-stream networks are then applied to the short snippets to obtain the initial action class prediction scores. The original TSN finally uses a segmental consensus function to combine the outputs from multiple short snippets to predict the action class probabilities for the video as a whole. Here, motivated by \cite{diba_tle_2016}, we encode the features from different segments through compact bilinear models \cite{compact_bilinear_cvpr2016} as shown in Figure \ref{fig:network}.

\noindent \textbf{Spatial stream} A standard spatial stream takes a single video frame as input. Here, we extend this to multiple frames. Hence, our random temporal skipping also works for the spatial stream. 

\noindent \textbf{Temporal stream} A standard temporal stream takes a stack of 10 optical flow images as input. However, the pre-computation of optical flow is time consuming, storage demanding and sub-optimal for action recognition. Motivated by \cite{hidden_zhu_18}, we propose to use a CNN to learn optical flow from video frames and directly feed the predictions to the temporal stream. We name this optical flow CNN MotionNet as shown in Figure \ref{fig:network}. 

For the MotionNet, we treat optical flow estimation as an image reconstruction problem \cite{densenet_flow_icip17,dilate_flow_icip18}. The intuition is that if we can use the predicted flow and the next frame to reconstruct the previous frame, our model has learned a useful representation of the underlying motion. Suppose we have two consecutive frames $I_{1}$ and $I_{2}$. Let us denote the reconstructed previous frame as $I_{1}^{\prime}$. The goal then is to minimize the photometric error between the true previous frame $I_{1}$ and the reconstructed previous frame $I_{1}^{\prime}$: 
\begin{equation}
	L_{\text{reconst}} = \frac{1}{N} \sum_{i, j}^{N} \rho ( I_{1}(i, j) - I_{1}^{\prime}(i,j) ).
	\label{eq:reconstruction_loss}
\end{equation}
$N$ is the number of pixels. The reconstructed previous frame is computed from the true next frame using inverse warping, $I_{1}^{\prime}(i,j) = I_{2}(i+U_{i,j}, j+V_{i,j})$, accomplished through spatial transformer modules \cite{stn_nips15} inside the CNN. $U$ and $V$ are the horizontal and vertical components of predicted optical flow. We use a robust convex error function, the generalized Charbonnier penalty $\rho(x) = (x^{2} + \epsilon^{2})^{\alpha}$, to reduce the influence of outliers. $\alpha$ is set to $0.45$.

However, \cite{hidden_zhu_18} is based on a simple brightness constancy assumption and does not incorporate reasoning about occlusion. This leads to noisier motion in the background and inconsistent flow around human boundaries. As we know, motion boundaries are important for human action recognition. Hence, we extend \cite{hidden_zhu_18} by incorporating occlusion reasoning, hoping to learn better flow maps for action recognition.

In particular, our unsupervised learning framework should not employ the brightness constancy assumption to compute the loss when there is occlusion. Pixels that become occluded in the second frame should not contribute to the photometric error between the true and reconstructed first frames in Equation \ref{eq:reconstruction_loss}.
We therefore mask occluded pixels when computing the image reconstruction loss in order to avoid learning incorrect deformations to fill the occluded locations. Our occlusion detection is based on a forward-backward consistency assumption. That is, for non-occluded pixels, the forward flow should be the inverse of the backward flow at the corresponding pixel in the second frame. We mark pixels as being occluded whenever the mismatch between these two flows is too large. 
Thus, for occlusion in the forward direction, we define the occlusion flag $o^{f}$ be 1 whenever the constraint
\begin{equation}
	|M^{f} + M^{b}_{M^{f}}|^{2} < \alpha_{1} \cdot (|M^{f}|^{2} + |M^{b}_{M^{f}}|^{2}) + \alpha_{2}
	\label{eq:occlusion}
\end{equation}
is violated, and 0 otherwise. $o^{b}$ is defined in the same way, and $M^{f}$ and $M^{b}$ represent forward and backward flow. We set $\alpha_{1}$=0.01, $\alpha_{2}$=0.5 in all our experiments. 
Finally, the resulting occlusion-aware loss is represented as:
\begin{equation}
	L = (1 - o^{f}) \cdot L_{\text{reconst}}^{f} + (1 - o^{b}) \cdot L_{\text{reconst}}^{b}
	\label{eq:data_loss}
\end{equation}

Once we learn a geometry-aware MotionNet to predict motions between consecutive frames, we can directly stack it to the original temporal CNN for action mapping. Hence, our whole temporal stream is now end-to-end optimized without the computational burden of calculating optical flow. 

\subsection{Compact Bilinear Encoding}

In order to learn a compact feature for an entire video, we need to aggregate information from different segments. There are many ways to accomplish this goal, such as taking the maximum or average, bilinear pooling, Fisher Vector (FV) encoding \cite{FisherVector}, etc. Here, we choose compact bilinear pooling \cite{compact_bilinear_cvpr2016,Miao_2018_CVPR,Miao_2018_AAAI} due to its simplicity and good performance. 

The classic bilinear model computes a global descriptor by calculating:
\begin{equation}
	B = \phi (F \otimes F^{'}).
	\label{eq:bilinear}
\end{equation}
Here, $F$ are the feature maps from all channels in a specific layer, $\otimes$ denotes the outer product, $\phi$ is the model parameters we are going to learn and B is the bilinear feature. However, due to the many channels of feature maps and their large spatial resolution, the outer product will result in a prohibitively high dimensional feature representation.

For this reason, we use the Tensor Sketch algorithm as in \cite{compact_bilinear_cvpr2016} to avoid the computational intensive outer product by an approximate projection. Such approximation requires almost no parameter memory. We refer the readers to \cite{compact_bilinear_cvpr2016} for a detailed algorithm description.

After the approximate projection, we have compact bilinear features with very low feature dimension. Compact bilinear pooling can also significantly reduce the number of CNN model parameters since it can replace fully-connected layers, thus leading to less over-fitting. We will compare compact bilinear pooling to other feature encoding methods in later sections.

\subsection{Spatio-Temporal Fusion}

Following the testing scheme of \cite{twostream2014,wanggoodpractice2015,xue2018deep}, we evenly sample $25$ frames/clips for each video. For each frame/clip, we perform $10$x data augmentation by cropping the $4$ corners and $1$ center, flipping them horizontally and averaging the prediction scores (before softmax operation) over all crops of the samples. In the end, we obtain two predictions, one from each stream. We simply late fuse them by weighted averaging. The overview of our framework is shown in Figure \ref{fig:network}. 

\section{Experiments}

\paragraph{\bf Implementation Details} 

For the CNNs, we use the Caffe toolbox \cite{jia2014caffe}. Our MotionNet is first pre-trained using Adam optimization with the default parameter values. It is a 25 layer CNN with an encoder-decoder architecture \cite{hidden_zhu_18}. The initial learning rate is set to $3.2\times10^{-5}$ and is divided in half every $100$k iterations. We end our training at $400$k iterations. Once MotionNet can estimate decent optical flow, we stack it to a temporal CNN for action prediction. Both the spatial CNN and the temporal CNN are BN-Inception networks pre-trained on ImageNet challenges \cite{imagenet_cvpr09}. We use stochastic gradient descent to train the networks, with a batch size of $128$ and momentum of $0.9$.  We also use horizontal flipping, corner cropping and multi-scale cropping as data augmentation. Take UCF101 as an example. For the spatial stream CNN, the initial learning rate is set to $0.001$, and divided by $10$ every $4$K iterations. We stop the training at $10$K iterations.  For the stacked temporal stream CNN, we set different initial learning rates for MotionNet and the temporal CNN, which are $10^{-6}$ and $10^{-3}$, respectively. Then we divide the learning rates by $10$ after $5$K and $10$K. The maximum iteration is set to $16$K. Other datasets have the same learning process except the training iterations are different depending on the dataset size. 

\subsection{Trimmed Video}

\paragraph{\bf Dataset} 
In this section, we adopt three trimmed video datasets to evaluate our proposed method, UCF101 \cite{ucf101}, HMDB51 \cite{hmdb51} and Kinetics \cite{kinetics}.
UCF101 is composed of realistic action videos from YouTube. It contains $13,320$ video clips distributed among $101$ action classes. HMDB51 includes $6,766$ video clips of $51$ actions extracted from a wide range of sources, such as online videos and movies.
Both UCF101 and HMDB51 have a standard three-split evaluation protocol and we report the average recognition accuracies over the three splits. 
Kinetics is similar to UCF101, but substantially larger. It consists of approximately $400,000$ video clips, and covers $400$ human action classes. 

\begin{table}[t]
	\begin{center}
		\caption{Necessity of multirate analysis. RTS indicates random temporal skipping. Fixed sampling means we sample the video frames by a fixed length (numbers in the brackets, e.g., 1, 3, 5 frames apart). Random sampling indicates we sample the video frames by a random length of frames apart.  \label{tab:necessity}}
			\begin{tabular}{ c | c | c }
				\hline
				Method			         			&    without RTS    &    with RTS         \\
				\hline		
				\hline
				No Sampling   	 						&   $95.6$  &   $96.4$ 	      	\\
				\hline
				Fixed Sampling (1)  	 				&   $93.4$  &   $95.8$ 	    \\
				\hline
				Fixed Sampling (3)    	 				&   $91.5$  &   $94.9$ 	       	\\
				\hline
				Fixed Sampling (5)     	 				&   $88.7$  &   $92.3$ 	     	\\
				\hline
				Random Sampling  	 				&   $87.0$  &   $92.3$ 	    	\\
				\hline
			\end{tabular}
	\end{center}
\end{table} 

\paragraph{\bf Necessity of Multirate Analysis} 
First, we demonstrate the importance of multirate video analysis. We use UCF101 as the evaluation dataset. We show that a well-trained model with a fixed frame-rate does not work well when the frame-rate differs during testing. 
As shown in Table \ref{tab:necessity}, no sampling means the dataset does not change. Fixed sampling means we manually sample the video frames by a fixed length (numbers in the brackets, e.g., 1, 3, 5 frames apart). Random sampling indicates we manually sample the video frames by a random length of frames apart. We set the maximum temporal stride to 5. ``with RTS'' and ``without RTS'' indicates the use of our proposed random temporal skipping strategy during model training or not. Here, all the samplings are performed for test videos, not training videos. This is used to simulate frame-rate changes between the source and target domains.

We make several observations. First, if we compare the left and right columns in Table \ref{tab:necessity}, we can clearly see the advantage of using random temporal skipping and the importance of multirate analysis. Without RTS, the test accuracies are reduced dramatically when the frame-rate differs between the training and test videos. When RTS is adopted, the performance decrease becomes much less significant. 
Models with RTS perform $5\%$ better than those without RTS on random sampling (last row). 
Second, in the situation that no sampling is performed (first row in Table \ref{tab:necessity}), models with RTS perform better than those without RTS. This is because RTS helps to capture more temporal variation. It helps to regularize the model during training, acting like additional data augmentation.
Third, if we change fixed sampling to random sampling (last two rows in Table \ref{tab:necessity}), we can see that the recognition accuracy without RTS drops again, but the accuracy with RTS remains the same. This demonstrates that our proposed random temporal skipping captures frame-rate invariant features for human action recognition. 

One interesting thing to note is that, with the increase of sampling rate, the performance of both approaches decrease. This maybe counter-intuitive because RTS should be able to handle the varying frame-rate. The reason for lower accuracy even when RTS is turned on is because videos in UCF101 are usually short. Hence, we do not have as many training samples with large sampling rates as those with small sampling rates. We will show in the next section that when the videos are longer, models with RTS can be trained better. 

\begin{figure}[t]
	\centering
	\includegraphics[width=1.0\linewidth]{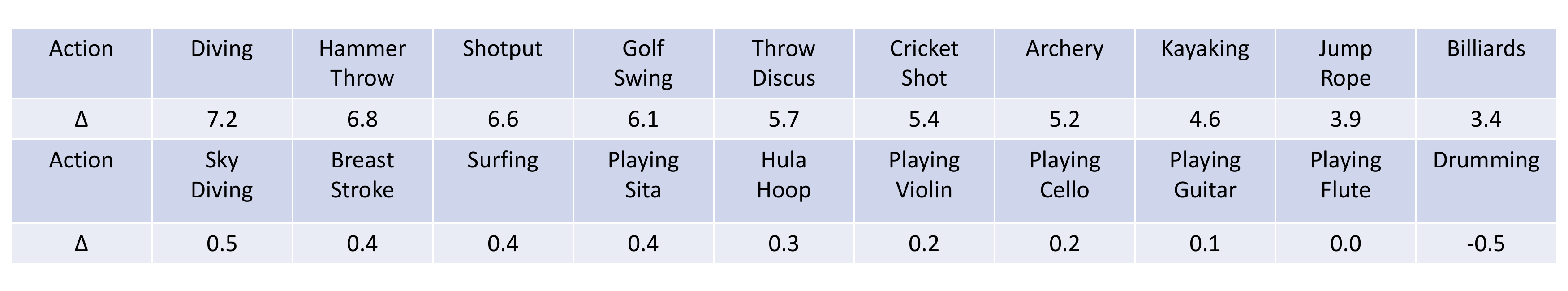}
	\caption{Top-10 classes that benefit most (top) and least (bottom) in UCF101}
	\label{fig:perclass}
\end{figure}


\paragraph{\bf Per-Class Breakdown} 
Here, we perform a per-class accuracy breakdown to obtain insights into why random temporal skipping works and how it helps. We choose the results from the last row in Table \ref{tab:necessity} to compare. 

We list, in Figure \ref{fig:perclass} below, the 10 classes in UCF101 that benefit the most from RTS and the 10 that benefit the least. The actions that benefit the most tend to exhibit varying motion speeds. 
The actions that benefit the least can either be considered still, and can thus be recognized by individual frames regardless of how they are sampled, or considered repetitive, and so a constant sampling rate is sufficient. Hence, our proposed random temporal skipping effectively handles different motion speeds.



\begin{table}[t]
	\begin{center}
		\caption{Comparison with various feature aggregation methods on UCF101 and HMDB51.  Compact bilinear pooling achieves the best performance in terms of classification accuracy.  \label{tab:encoding}}
			\begin{tabular}{ c | c | c }
				\hline
				Method			         			&    UCF101    &    HMDB51        \\
				\hline		
				\hline
				FC   	 									&   $94.9$  &   $69.7$ 	      	\\
				\hline
				BoVW  	 								&   $92.1$  &   $65.3$ 	    \\
				\hline
				VLAD   	 								&   $94.3$  &   $66.8$ 	       	\\
				\hline
				FV     	 									&   $93.9$  &   $67.4$ 	     	\\
				\hline
				Compact Bilinear Pooling  	 				&   $\mathbf{96.4}$  &   $\mathbf{72.5}$ 	    	\\
				\hline
			\end{tabular}
	\end{center}
\end{table} 

\paragraph{\bf Encoding Methods Comparison} 

In this section, we compare different feature encoding methods and show the effectiveness of compact bilinear encoding. In particular, we choose four widely adopted encoding approaches: Bag of Visual Words (BoVW), Vector of Locally Aggregated Descriptors (VLAD), Fisher Vector (FV) and Fully-Connected pooling (FC). 

FC is the most widely adopted feature aggregation method in deep learning era, thus will be served as baseline. We put it between the last convolutional layer and the classification layer, and set its dimension to 4096. FC will be learned end-to-end during training. BoVW, VLAD and FV are clustering based methods. Although there are recent attempts to integrate them into CNN framework \cite{dovf_lan_2017}, for simplicity, we do not use them in an end-to-end network. We first extract features from a pre-trained model, and then encode the local features into global features by one of the above methods. Finally, we use support vector machines (SVM) to do the classification. To be specific, suppose we have N local features, BoVW quantizes each of the N local features as one of k codewords using a codebook generated through k-means clustering. VLAD is similar to BoVW but encodes the distance between each of the N local features and the assigned codewords. FV models the distribution of the local features using a Gaussian mixture model (GMM) with k components and computes the mean and standard deviation of the weighted difference between the N local features and these k components. In our experiments, we project each local feature into 256 dimensions using PCA and set the number of clusters (k) as 256. This is similar to what is suggested in \cite{LCDXu2015} except we do not break the local features into multiple sub-features. For the bilinear models, we retain the convolutional layers of each network without the fully-connected layers. The convolutional feature maps extracted from the last convolutional layers (after the rectified activation) are fed as input into the bilinear models. Here, the convolutional feature maps for the last layer of BN-Inception produces an output of size 14 $\times$ 14 $\times$ 1024, leading to bilinear features of size 1024 $\times$ 1024, and 8,196 features for compact bilinear models. 

As can be seen in Table \ref{tab:encoding}, our compact bilinear encoding achieves the best overall performance (two-stream network
results). This observation is consistent with \cite{diba_tle_2016}. It is interesting that the more complicated encoding methods, BoVW, FV and VLAD, all perform much worse than baseline FC and compact bilinear pooling. We conjecture that this is because they are not end-to-end optimized. 

\begin{figure}[t]
	\centering
	\includegraphics[width=1.0\linewidth]{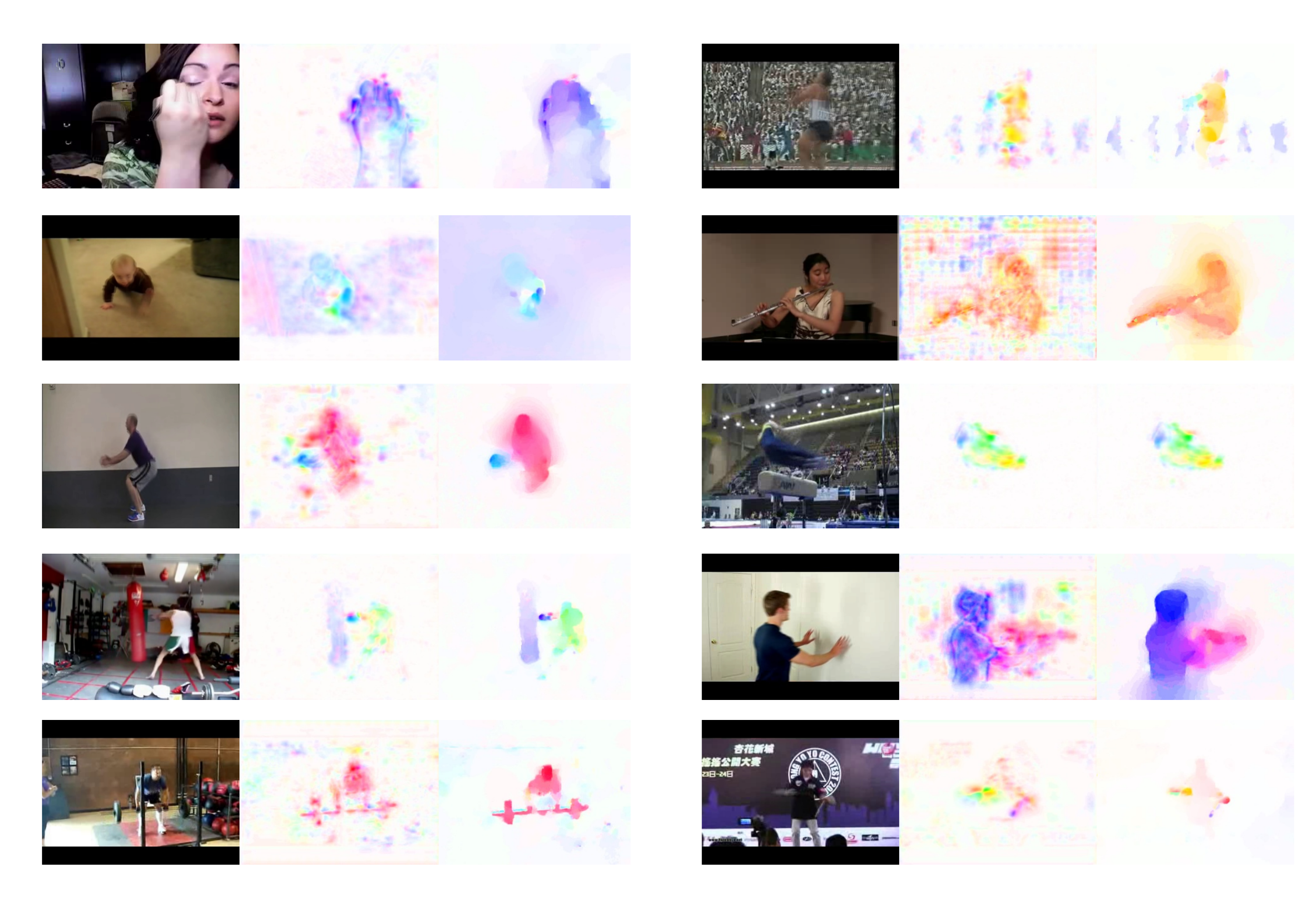}
	\caption{Sample visualizations of UCF101 dataset to show the impact of reasoning occlusion during optical flow estimation. Left: overlapped image pairs. Middle: MotionNet without occlusion reasoning. Right: MotionNet with occlusion reasoning. The figure is best viewed in color. We can observe the clear improvement brought by occlusion reasoning. }
	\label{fig:actionflow}
\end{figure}

\paragraph{\bf Importance of Occlusion-Aware} 

One of our contributions in this work is introducing occlusion reasoning into the MotionNet \cite{hidden_zhu_18} framework. Here, we show sample visualizations to demonstrate its effectiveness.  

As can be seen in Figure \ref{fig:actionflow}, optical flow estimates with occlusion reasoning are much better than those without. Occlusion reasoning can remove the background noise brought by invalid brightness constancy assumptions, reduce checkerboard artifacts, and generate flows with sharper boundaries due to awareness of disocclusion. 
Quantitatively, we use these two flow estimates as input to the temporal stream. Our network with occlusion reasoning performs $0.9\%$ better than the baseline \cite{hidden_zhu_18} on UCF101 (95.5 $\rightarrow$ 96.4). This makes sense because a clean background of optical flow should make it easier for the model to recognize the action itself than the context. We show that we can obtain both  better optical flow and higher accuracy in action recognition by incorporating occlusion reasoning in an end-to-end network.

\subsection{Untrimmed Video}

\paragraph{\bf Dataset} 
In this section, we adopt three untrimmed video datasets to evaluate our proposed method, ActivityNet \cite{activityNet}, VIRAT 1.0 \cite{virat_cvpr2011} and VIRAT 2.0 \cite{virat_cvpr2011}. 
For ActivityNet, we use version 1.2 which has 100 action classes. Following the standard evaluation split, 4,819 training and 2,383 validation videos are used for training and 2,480 videos for testing. 
VIRAT 1.0 is a surveillance video dataset recorded in different scenes. Each video clip contains 1 to 20 instances of activities from 6 categories of person-vehicle interaction events including: loading an object to a vehicle, unloading an object from a vehicle, opening a vehicle trunk, closing a vehicle trunk, getting into a vehicle, and getting out of a vehicle.
VIRAT 2.0 is an extended version of VIRAT 1.0. It includes 5 more events captured in more scenes: gesturing, carrying an object, running, entering a facility and exiting a facility. We follow the standard train/test split to report the performance. 

\begin{figure*}[t]
	\centering
	\includegraphics[width=1.0\linewidth]{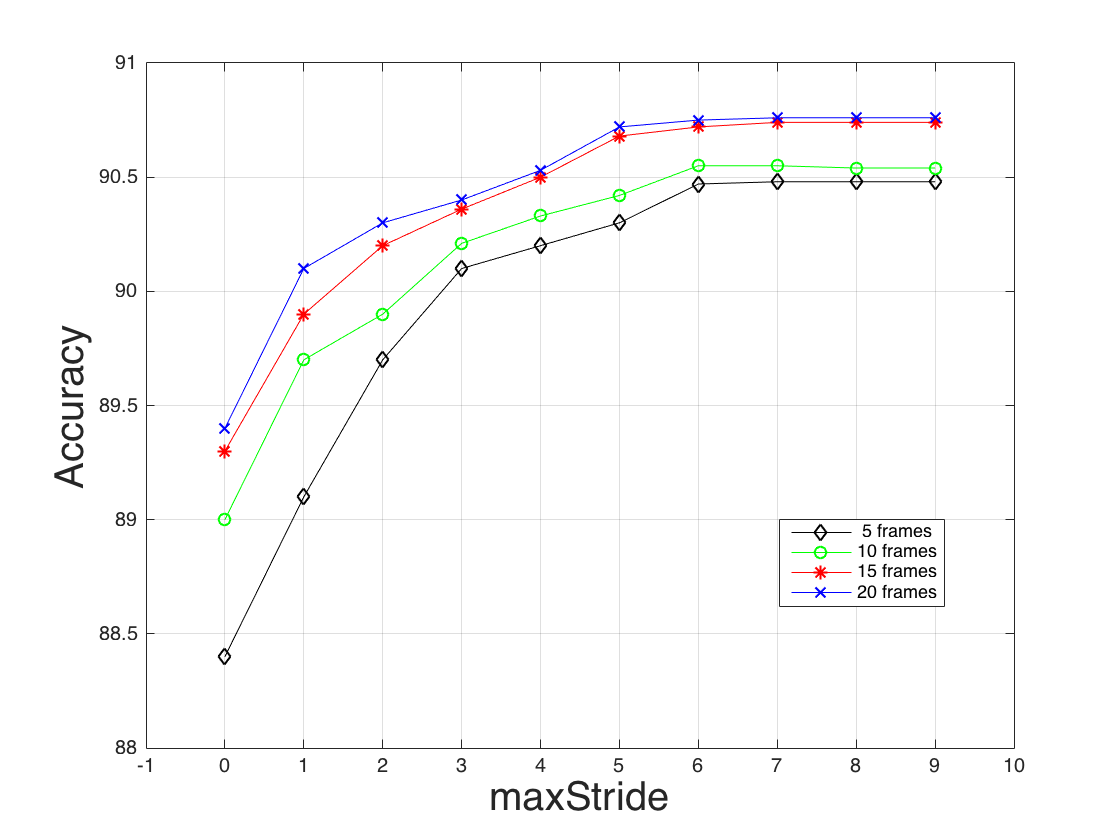}
	\caption{Action recognition accuracy on ActivityNet. We observe that the longer temporal context we utilize, the better performance we obtain. }
	\label{fig:rts}
\end{figure*}


\paragraph{\bf Investigate Longer Temporal Context} 

In the previous section, we demonstrated that a well-trained model with a fixed frame-rate does not work well when frame-rate differs during testing. Here, we show that using a longer temporal context by random temporal skipping is useful for action recognition. We use ActivityNet as the evaluation dataset because most videos in ActivityNet are long (5 to 10 minutes) so that we can explore more speed variations.

Recall from Equation \ref{eq:skip} that \textit{maxStride} is a threshold value indicating the maximum distance we can skip in the temporal domain. We set it from 0 frames to 9 frames apart, indicating no sampling to the longest temporal coverage. As shown in Figure \ref{fig:rts}, we can see that the longer temporal context we utilize, the higher action recognition accuracy we obtain. One interesting observation is that the performance starts to saturate when \textit{maxStride} is equal to 6. After that, longer temporal context does not help much. We think this may be due to the fact that the CNNs can not capture the transitions between frames that are so far away.

In addition, we investigate the impact of the number of sampled frames. We choose 5, 10, 15 and 20 frames as the length of the input video clip. As we can see in Figure \ref{fig:rts}, more sampled frames always improves the action recognition accuracy. This demonstrates that longer temporal information benefits video understanding. With 20 input frames and a \textit{maxStride} of 6, our method can have a temporal coverage of over 120 fames, which is about 5 seconds. Such a time duration is enough for analyzing most actions or events. For UCF101 and HMDB51 datasets, 5 seconds can cover the entire video.

\begin{table*}[t]
	\begin{center}
		\caption{Comparison to state-of-the-art approaches in accuracy ($\%$). \label{tab:sota}}
		\begin{tabular}{  c | c | c | c | c | c | c}
			\hline
			Method																	&    UCF101  &    HMDB51   & Kinetics &   ActivityNet  &   VIRAT 1.0   &  VIRAT 2.0   \\
			\hline		
			\hline
			Two-Stream \cite{twostream2014}					&   $88.0$ 	&    $59.4$	&   $62.2$ 		&   $71.9$ 	&    $80.4$		&   $92.6$ 	\\	
			C3D  \cite{c3d2015}										&   $82.3$ 	&    $49.7$		&   $56.1$ 	&   $74.1$ 	&    $75.8$		&   $87.5$ 	\\	
			TDD  \cite{tddwang2015}								&   $90.3$ 	&    $63.2$		&   $-$ 	&   $-$ 	&    $86.6$		&   $93.2$ 	\\	
			LTC \cite{longTemporalConv2016}					&   $91.7$ 	&    $64.8$		&   $-$ 	&   $-$ 	&    $-$		&   $-$ 	\\	
			Depth2Action \cite{depth2action}					&   $93.0$ 	&    $68.2$		&   $68.7$ 	&   $78.1$ 	&    $89.7$		&   $94.1$ 	\\	
			TSN \cite{TSN2016}										&   $94.0$ 	&    $68.5$		&   $73.9$ 	&   $89.0$ 	&    $-$		&   $-$ 	\\	
			TLE \cite{diba_tle_2016}								&   $95.6$ 	&    $71.1$		&   $75.6$ 	&   $-$ 	&    $-$		&   $-$ 	\\	
			\hline
			\hline
			Ours				&   $\mathbf{96.4}$ 	&    $\mathbf{72.5}$		&   $\mathbf{77.0}$ 	&   $\mathbf{91.1}$ 	&    $\mathbf{94.2}$		&   $\mathbf{97.1}$ 	\\	
			\hline
		\end{tabular}
	\end{center}
\end{table*} 

\subsection{Comparison to State-of-the-Art}
We compare our method to recent state-of-the-art on the six video benchmarks. As shown in Table \ref{tab:sota}, our proposed random temporal skipping is an effective data augmentation technique, which leads to the top performance on all evaluation datasets. 

For the trimmed video datasets, we obtain performance improvements of 0.8$\%$ on UCF101, 1.4$\%$ on HMDB51 and 1.4$\%$ on Kinetics. Because the videos are trimmed and short, we do not benefit much from learning longer temporal information. The improvement
for UCF101 is smaller as the accuracy is already saturated on this dataset. Yet, our simple random temporal skipping strategy can improve it further. 

For the three untrimmed video datasets, we obtain significant improvements, 1.8$\%$ on ActivityNet, 4.5$\%$ on VIRAT 1.0 and 3.0$\%$ on VIRAT 2.0. This demonstrates the importance of multirate video analysis in complex real-world applications, and the effectiveness of our method. We could adapt our approach to real-time action localization due to the precise temporal boundary modeling.

There is a recent work I3D \cite{I3D_Carreira_cvpr17} that reports higher accuracy on UCF101 (98.0$\%$) and HMDB51 (80.7$\%$). However, it uses additional training data (\cite{kinetics}) and the network is substantially deeper, which is not a fair comparison to the above approaches. In addition, we would like to note that our approach is real-time because no pre-computation of optical flow is needed. We are only about $1\%$ worse than I3D, but 14 times faster. 

\section{Conclusion}
In this work, we propose a simple yet effective strategy, termed random temporal skipping, to handle multirate videos. It can benefit the analysis of long untrimmed videos by capturing longer temporal contexts, and of short trimmed videos by providing extra temporal augmentation. The trained model using random temporal skipping is robust during inference time. We can use just one model to handle multiple frame-rates without further fine-tuning. 
We also introduce an occlusion-aware CNN to estimate better optical flow for action recognition on-the-fly. Our network can run in real-time and obtain state-of-the-art performance on six large-scale video benchmarks. 

In the future, we would like to improve our framework in several directions. First, due to the inability of CNNs to learn large motions between distant frames, we will incorporate recurrent neural networks into our framework to handle even longer temporal contexts. Second, we will apply our method to online event detection since our model has a good trade-off between efficiency and accuracy. Third, we will study the fusion of two streams and compare to recent spatiotemporal feature learning work \cite{spatiotemporal_xie,url_cvpr18}.

\noindent \textbf{Acknowledgement} We gratefully acknowledge the support of NVIDIA Corporation through the donation of the Titan Xp GPUs used in this work.

	%
	%
	%
	\bibliographystyle{splncs04}
	\bibliography{egbib}
	
\end{document}